\newcommand{\dataset}{{\cal D}}

\documentclass[sn-mathphys,Numbered]{sn-jnl}
\usepackage{natbib}
\usepackage{graphicx}%
\usepackage{multirow}%
\usepackage{amsmath,amssymb,amsfonts}%
\usepackage{amsthm}%
\usepackage{mathrsfs}%
\usepackage[title]{appendix}%
\usepackage{xcolor}%
\usepackage[dash,dot]{dashundergaps}
\usepackage{tablefootnote}
\usepackage{textcomp}%
\usepackage{manyfoot}%
\usepackage{booktabs}%
\usepackage{algorithm}%
\usepackage{algorithmicx}%
\usepackage{algpseudocode}%
\usepackage{listings}%

\theoremstyle{thmstyleone}

\theoremstyle{thmstyletwo}

\theoremstyle{thmstylethree}

\begin{document}
\renewcommand{\arraystretch}{1.5}

\title[Article Title]{Area-norm COBRA on Conditional Survival Prediction}

\author[1]{\fnm{Rahul} \sur{Goswami}}\email{rahul.goswami@iitg.ac.in}

\author[2, 3]{\fnm{Arabin Kr.} \sur{Dey}}\email{arabin.k.dey@gmail.com}

\affil*[1, 2]{\orgdiv{Department}, \orgname{of Mathematics}, \orgaddress{\street{IIT Guwahati}, \city{Guwahati}, \postcode{781039}, \state{Assam}, \country{India}}}
\affil*[3]{Corresponding Author : Arabin Kumar Dey; Email address : arabin.k.dey@gmail.com }

\abstract{The paper explores a different variation of combined regression strategy to calculate the conditional survival function. We use regression based weak learners to create the proposed ensemble technique. The proposed combined regression strategy uses proximity measure as area between two survival curves. The proposed model shows a construction which ensures that it performs better than the Random Survival Forest. The paper discusses a novel technique to select the most important variable in the combined regression setup. We perform a simulation study to show that our proposition for finding relevance of the variables works quite well. We also use three real-life datasets to illustrate the model.}


\maketitle

\section{Introduction}\label{sec1}

  Prediction of Conditional survival function is an important relevant information for doctors, clinicians, and insurance companies. Function prediction has a large number of applications in many areas of science and technology. This is a challenging problem.  Researchers attempted to provide a reasonable generic solution over decades.  After the Cox proportional hazard model (\cite{cox1975partial}, \cite{cox1972regression}), one of the breakthrough contributions to this problem is Random Survival Forest by Ishawaran \cite{ishwaran2008random}.  The solution encourages researchers to explore the problem through different other neural network models [\cite{faraggi1995neural}, \cite{katzman2018deepsurv}, \cite{faraggi1995neural,katzman2018deepsurv}] and ensemble techniques which include XGboost, Adaboost (\cite{bellot2018boosted}) etc. In technical language, this is a problem of constructing a functional regression. Goswami et al (\cite{goswami2022concordance}, \cite{goswami2022integrated}) provide different solutions to this problem in a combined regression setup.  However, the solution has issues and demands a more detailed exploration to get a complete idea of the different possible implementations and their benefits.  The paper addresses a different variation of the COBRA which works for certain aspects and certain types of datasets. The paper proposes to use a different proximity measure for this combined regression strategy and a specific set of weak learners which ensures that the proposed algorithm works much better than the usual Random Survival Forest. It ensures the algorithm will work at least as well as a random survival forest if the dataset consists of the censored observation up to a certain level.   

 All implementation in this paper assumes datasets are sufficiently large. This ensures the proximity measure to include a reasonable number of points in the prediction set. It uses Kaplan Meier estimation (\cite{kaplan1958nonparametric}) or Nelson-Aalen \cite{aalen1978nonparametric,nelson1972theory} estimator to estimate the survival function or cumulative hazard function respectively based on all similar observations.  We use both concordance measure and integrated Brier score in a censored setup to tune the hyperparameters of the model and use them to understand the goodness of fit for each dataset used in this paper. The approach does not use similar weak learners as used by Goswami et al. The weak learner's choices in this paper follow the same as the original combined regression paper with its similar extension in functional regression setup. Thus we take each base learner as a survival predictor.   We use different survival predictors instead of just one type of survival predictor used by Goswami et al. : (1) Cox model (ridge) \cite{simon2011regularization} (2) Cox model (lasso) \cite{tibshirani1997lasso} (3) random survival forest \cite{breiman2001random} (4) KNN survival (\cite{chen2019nearest}, \cite{beran1981nonparametric} ) (5) Survival Tree \cite{breiman2001random}. These are exact functional extensions of the following models (1) Ridge (2) Lasso (3) Random Forest \cite{breiman2001random} (4) KNN \cite{fix1952discriminatory} (5) Decision tree used respectively in original cobra for usual regression.  We observe the choice can help us to ensure that the algorithm works better than a Random survival forest for sufficiently large datasets with lesser censored data. The experimental results open a large number of issues and questions.  We address the issues in other research articles.
  
  The relevance of the covariates is also an important aspect of survival analysis. Earlier statistical models like Cox Proportional Hazard Model \cite{cox1972regression} and Accelerated Failure Time Model \cite{kalbfleisch2011statistical} have a clear idea of the covariates, which are relevant for the happening of the event of interest at a specific time.  The importance of the covariates in predicting the survival function measures the relevance of the covariates. Random Survival Forest \cite{ishwaran2008random} uses the out-of-bag error (\cite{breiman2001random}, \cite{ishwaran2008random}, \cite{ishwaran2007variable}, \cite{gromping2009variable}, \cite{genuer2010variable}) and minimal depth procedure.  Feature importance is crucial for providing the interpretability of the model. Independent approaches like LIME \cite{ribeiro2016should} and Shapley Value \cite{strumbelj2010efficient} use variable importance concepts in ordinary regression setup.  The approaches like SurvLIME (\cite{kovalev2020survlime, kovalev2020robust}, \cite{utkin2020survlime}) and SurvSHAP (\cite{alabdallah2022survshap}, \cite{krzyzinski2023survshap}) serves the similar purpose of bringing interpretability of survival model through measuring the importance of features.
  
 In this paper, we propose a novel technique to find out the most important variable. It can also find a score to rank the importance.  The technique exploits the cobra structure to find the variable importance. The proposition has support on a simulated framework.  

 The organization of the paper is as follows.  Section \ref{sec2} provides a brief overview of the notations and basic framework of survival analysis and original combined regressions strategy used for regression.  We describe the proposed framework for aggregating the survival-based functional models in Section \ref{sec3}. Numerical results of the experiments conducted to evaluate the performance of the proposed framework are available in Section \ref{sec4}.  A novel proposition for ranking the variables or selecting the most important variable exploiting the Cobra Structure is available in Section \ref{sec5}. Finally, Section \ref{sec6} concludes the paper and discusses future research directions.
 
\section{Background}\label{sec2}

\subsection*{Important Notations and Technical Specification of the Problem}

  Our goal is to develop individual survival function. Each patient $i$ have a d-dimensional 
covariate $x_i \in \mathcal{X}$ which represents the unique characteristics of the patient.
Here, $\mathcal{X}$ is the covariate space. An outcome of interest $T_i \in \mathbb{R}^{+}$, 
representing the time of occurrence of the event of interest.  The patient may drop out of the study before the event of interest occurs. We denote this dropout time or censored time as $C_i \in \mathbb{R}^{+}$.  
The observations carry another variable, $\delta_i \in \{0,1\}$. $\delta_i = 1$ when event of interest is available, and $\delta_i = 0$ if the event of
interest is not available. 

We can write $S(t \mid x_i) $ as  
\begin{equation}
 S(t \mid x_i) = \mathbb{P}(T_i > t \mid x_i)
\end{equation}

It represents the probability of survival of the patient $i$ beyond time $t$ given the covariate $x_i$.
Note that, our aim is to estimate the survival function $S$ specific to the patient $i$, not the unconditional survival function..
The relationship between the patient covariates and survival function is to be estimated
by a dataset $\dataset_n$ comprising of $n$ individuals assumed to be drawn from the random 
tuple $(X_i, \delta_i T_i + (1 - \delta_i)C_i , \delta_i)$ . For simplicity we denote the
random tuple $(X_i, \delta_i T_i + (1 - \delta_i)C_i , \delta_i)$ by $(X_i, Y_i, \delta_i)$.
Since, we do not observe both $T_i$ and $C_i$ concurrently in practice, we denote the observed
time by $Y_i = \delta_i T_i + (1 - \delta_i)C_i$
 
\subsection*{Combined Regression Strategy for Usual Regression Framework}

Let's consider the regression setting, where $\dataset_n = (X_1 , Y_1), \cdots, (X_n , Y_n)$ are independent
and identically distributed samples from $(\mathbf{X},Y) \in \mathbb{R}^p \times \mathbb{R}$. We assume
$\mathbb{E}Y^2 < \infty$. In a typical regression setup we are interested in estimating the conditional
expectation of $ Y $ given $ X=x $ i.e $ E(Y \mid X=x) $.

\noindent In COBRA setup, we split the the dataset $\dataset_n$ into two parts, one part $\dataset_k = (X_1 , Y_1) \cdots (X_k , Y_k)$
and the other part $\dataset_{l} = (X_{k+1} , Y_{k+1}), \cdots, (X_n , Y_n)$ where $l = n - k \geq 1$. We denote $\dataset_{l} = (X_1 , Y_1), \cdots, (X_l , Y_l)$.

 Suppose, we have M competing estimators, referred to as \textit{machines} in COBRA setup. Each machine $m$ is
trained on $\dataset_k$ and referred as $r_{k1}, r_{k2}, \cdots, r_{kM}$, here we note that $r_{ki}$ is a
machine trained only on $\dataset_k$ and capable of estimating $E(Y \mid X=x)$ for any $ x \in \mathbb{R}^p $.
Then the estimate of $E(Y \mid X=x)$ by COBRA is given by

\begin{equation}
  \hat{E}_{COBRA}(Y \mid X=x) =  \sum_{i=1}^{l} W_{n, i}(x) Y_i
\end{equation}

where $W_{n, i}(x)$ is the weight of the $i^{th}$ sample in $\dataset_{l}$ and is given by
\begin{equation}
  W_{n, i}(x) = \frac{\mathbb{I}(\cap_{m=1}^M \mid r_{k,m}(X_i) - r_{k,m}(x)\mid \leq \epsilon)}{\sum_{j=1}^l \mathbb{I}(\cap_{m=1}^M \mid r_{k,m}(X_j) - r_{k,m}(x) \mid \leq \epsilon)}
\end{equation}

where, $\epsilon$ is a user specified parameter. The intuition behind the weight is that if the $i^{th}$
sample is close to the query point $x$ in all the machines, then the weight of the $i^{th}$ sample is
1 and if the $i^{th}$ sample is far from the query point $x$ in any machine, then the weight of
the $i^{th}$ sample is 0. This condition for consensus of a fraction of the machines brings a new parameter $\alpha$, which tells us how many machines should agree on the closeness
of the $i^{th}$ sample to the query point $x$.The weight of the $i^{th}$ sample is given by

\begin{equation}
 W_{n, i}(x) = \frac{\mathbb{I}(\sum_{m=1}^M \mathbb{I}( \mid r_{k,m}(X_i) - r_{k,m}(x) \mid \leq \epsilon) \geq M \alpha)}{\sum_{j=1}^l \mathbb{I}(\sum_{m=1}^M \mathbb{I}( \mid r_{k,m}(X_j) - r_{k,m}(x) \mid \leq \epsilon) \geq M \alpha)}
\end{equation}

COBRA gives a good approximation of the conditional expectation of $Y$ given $X=x$, there
is both theoretical and experimental proof that COBRA perform better that individual machine
in terms of quadratic loss.

\section{Proposed Algorithm for Functional Regression}\label{sec3}

  The structure of the survival data takes the shape of $\dataset_n : \{X_i, Y_i, \delta_i\}_{i=1}^{n}$, 
where $X_i$ signifies the covariate under consideration. Correspondingly, $Y_i$ represents the lesser 
of the event of interest occurrence time and the censoring time while $\delta_i$ serves as the indicator 
that offers insight into whether we observe the event of interest or not.

  Let us consider of the set survival models $\mathcal{M} = \{m_1, m_2 \dots m_{\mid \mathcal{M} \mid} \}$. 
Each model of the set $\mathcal{M}$ provides a survival function. 
Since we use five fold cross validation, we divide the dataset into five parts, where each part represents test set and rest as train set.   Suppose, we denote the train data as $\dataset_{n^{'}}$ which is of size $n^{'}$ and split the dataset $\dataset_{n^{'}}$ into $\dataset_k$ and $\dataset_l$ such that $k + l = n^{'}$. Let us assume the train and test dataset
as $\dataset_k = \{(X_i, Y_i, \delta_i)\}_{i=1}^k$ and $\dataset_l = \{(X_i, Y_i, \delta_i)\}_{i=k+1}^{n^{'}}$. 
We use the subdivision of the dataset $\dataset_k$ to train the base models, which splits out individual survival function for
a given query point $x$.   $S_{k, m}(t \mid x)$ denotes the estimated the survival function based on $m^{th}$ base model,
and $k$ is the dataset on which we train the base model. We define a function as follows:

\begin{equation}
  % label this equation
  \label{indicator1}
  \Gamma(x;X_j,\mathcal{M},\epsilon,\alpha) = \mathbb{I} \biggl( \sum_{m=1}^{\mid \mathcal{M} \mid} \mathbb{I}( d(S_{k,m}(t \mid x) , S_{k,m}(t \mid X_j))  \leq \epsilon) \geq \mid \mathcal{M} \mid\alpha \biggr)
\end{equation}

  The \textit{k} subscript notify that the prediction of the base models are trained on the dataset $\dataset_k$.
Here, $\mathbb{I}$ is the indicator function, which is 1 if the condition is true and 0 if the condition is false. $\epsilon$ is
the threshold distance, $\alpha$ is the fraction of base models in the in consensus of $\epsilon$ proximity of the query point $x$.
basically $\Gamma(.)$ is the indicator function, which is 1 if the $i^{th}$ individual is in the $(\epsilon , \alpha)-$proximity of the
query point $x$, otherwise 0.

In this paper we take the following area-norm to calculate the distance function :
\begin{equation}
  \label{distance}
  d(S_{i}(t),S_{j}(t)) = \frac{1}{t}\int_{0}^{t} \mid S_{i}(t) - S_{j}(t) \mid dt
\end{equation}

However, computing this area-norm is computationally expensive.  Let's assume $t_{0} = 0$.  We use the following Euler approximation as distance function between survival function :
    \begin{equation}
    \label{distance1}
      d(S_i(t),S_j(t)) = \frac{1}{t_{n^{'}} - t_1}\sum_{k=1}^{n^{'}} \mid S_i(t_k) - S_j(t_{k-1}) \mid (t_k - t_{k-1})
    \end{equation}

The other part of the dataset $\dataset_l$ aggregate the survival function of the base models.
We calculate the event time, event count and survivor count using the following equations,

\begin{equation}
  \label{cobraeventtime}
  \mathcal{Y}_{\Gamma}(x;h) = \biggl\{Y_j : j \geq k \cap \delta_j \Gamma(x;X_j,\mathcal{M},\epsilon,\alpha) = 1\biggr\}
\end{equation}

\begin{equation}
  \label{cobraeventcount}
  \mathcal{D}_{\Gamma}(t \mid x;h) = \sum_{j=k}^{n^{'}} \Gamma(x;X_j,\mathcal{M},\epsilon,\alpha)\delta_j\mathbb{I}(Y_j = t)
\end{equation}

\begin{equation}
  \label{cobrasurvivorcount}
  \mathcal{R}_{\Gamma}(t \mid x;h) = \sum_{j=k}^{n^{'}}\Gamma(x;X_j,\mathcal{M},\epsilon,\alpha)\mathbb{I}(Y_j \geq t)
\end{equation}

Then using the equations \ref{cobraeventtime}, \ref{cobraeventcount} and \ref{cobrasurvivorcount}, we can construct
the aggregated survival function $\hat{S}_{COBRA}(t \mid x;h)$ using the following equation

\begin{equation}
  \label{cobrakaplanmeier}
  \hat{S}_{COBRA}(t \mid x;h) = \prod_{t^{'} \in \mathcal{Y}_{\Gamma}(x;h)} \biggl( 1 - \frac{\mathcal{D}_{\Gamma}(t^{'} \mid x;h)}{\mathcal{R}_{\Gamma}(t^{'} \mid x;h)} \biggr)^{\mathbb{I}(t^{'} \leq t)}
\end{equation}

 The proposed algorithm is available in Algorithm \ref{algo1}.

% create algorithm
\begin{algorithm}[htbp]
  \caption{Overall flow of the proposed Algorithm}
  \label{algo1}
  \begin{algorithmic}[1]
    \Require
    \Statex $\dataset_n$ : Dataset of $n$ individuals
    \Statex $\dataset_{n^{'}}$ : Dataset of $n^{'}$ train individuals
    \Statex $\epsilon$ : Threshold distance
    \Statex $\alpha$ : Fraction of base models in the consensus of $\epsilon$ proximity of the query point $x$.
    \Statex $M$ : Number of base models
    \Ensure
    \Statex $\hat{S}(t \mid x)$ : Estimated survival function of the query point $x$ at time $t$
    \Statex
    \State Partition the whole data into five parts.  Take each part as test set.
    \State Split train data $\dataset_{n^{'}}$ into $\dataset_k$ and $\dataset_l$ such that $k + l = n^{'}$
    \State Train $M$ base models on $\dataset_k$ and predict the survival function $S_{k,m}(t \mid x)$ for $m = 1,2 \dots M$
    \State Calculate the distance $d(S_{k,m}(t \mid x), S_{k,m}(t \mid x_i))$ for $m = 1,2 \dots M$ and $i = k+1, k+2, \cdots, n^{'}$ using \ref{distance1}
    \State Calculate the indicator function $\Gamma_{i}(x)$ for $i = k+1, k+2, \cdots, n^{'}$ using \ref{indicator1}
    \State Calculate the estimated survival function $\hat{S}(t \mid x)$ using \ref{cobrakaplanmeier}
  \end{algorithmic}
\end{algorithm}

\section{Dataset Descriptions and Numerical Experiments}\label{sec4}

\subsection{Datasets}\label{subsubsec1}

We evaluate the performance of the proposed method on the following datasets:

\begin{itemize} 

\item \textbf{METABRIC} \footnote{Available at \url{https://github.com/chl8856/DeepHit}} : The Molecular Taxonomy of Breast Cancer International Con-
sortium (METABRIC) dataset comprises of gene expression profiles and clinical
attributes aimed at discerning breast cancer subtypes. Among the 1,981 patients
included, 888 (44.8\%) had complete follow-up data until mortality, while the
remaining 1,093 (55.2\%) were right-censored. Our analysis focuses on 21 publicly
accessible clinical features, including critical indicators such as tumor size and
lymph node involvement, with detailed descriptions available in Bilal et al. (2013) \cite{bilal2013improving}.
Missing values underwent imputation to ensure data integrity, with real-valued
features substituted by their mean and categorical features by their mode. We one-hot code the categorical 
variables, facilitating their integration into subsequent analytical methodologies. 

\item \textbf{FLCHAIN} \footnote{Available at \url{https://pypi.org/project/SurvSet/}} \cite{dispenzieri2012use, kyle2006prevalence} :   The FLCHAIN3 [\cite{dispenzieri2012use}, \cite{kyle2006prevalence}] dataset, introduced by Dispenzieri et al. \cite{dispenzieri2012use}, constitutes a publicly accessible resource designed to investigate the intricate association between serum free light chain levels and mortality. This dataset encompasses a diverse array of covariates, including age, gender, serum creatinine levels, and the presence of monoclonal gammopathy, all of which are pivotal factors in comprehending the dynamics of this relationship. This dataset is a critical resource for elucidating the multifaceted interplay between serum free light chains and pertinent covariates in the context of mortality studies.
  
  \item \textbf{RECID} \footnote{Available at \url{https://github.com/georgehc/npsurvival}} :   The Recidivism4 \cite{wooldridge2000recid} dataset is a publicly available dataset, widely utilised in criminology and social sciences, aimed at exploring the intricate dynamics of recidivism -- defined as the re-offending or re-incarceration of individuals previously involved with the criminal justice system. This dataset offers a comprehensive array of covariates essential for the analysis of recidivism, including demographic information such as age, gender, socio-economic factors and criminal history variables like prior convictions and offence type. This dataset serves as a fundamental resource for gaining insights into the multifaceted determinants of recidivism and contributes to informed policy and intervention strategies in criminal justice studies.
  
  \end{itemize}

% create a table with column name dataset , totalobservation , totalfeatures , %ofcensored , time-range
\begin{table}[ht]
  \centering
  \begin{tabular}{rrrrl}
    \hline
    Dataset & Total Observations & \#  Features$^{*}$ & \%  Censored & Five Number Summary (time) \\ 
    \hline
  metabric & 1981 &  79 & 55.2 & 3,1498,2632,4361,9218 \\ 
    flchain & 7874 &  10 & 72.5 & 0,2852,4302,4773,5215 \\ 
    recid & 1445 &  14 & 61.8 & 1,27,71,76,81 \\ 
     \hline
  \end{tabular}
  \footnotesize{$^*$ The total number of features  after preprocessing}
  \end{table}

\subsection{Experimental Setup}\label{subsubsec2}

 We conduct the following experiments to evaluate the performance of the proposed method. The experiment uses he five 
fold cross validation are reported for each dataset, \ref{tab:ibs} and \ref{tab:concordance}.  We tune all parameters for each dataset using hyperopt \cite{bergstra2013hyperopt}. The
parameters are given in table \ref{tab:parameters}.   Note that, we take objective function in hyperopt as Integrate Brier Score and negative Concordance Index.

% create a table with column name parameters and range tuning range
\begin{table}[htbp]
  \centering
  \caption{Parameters tuned with 1000 trials using hyperopt}
  \label{tab:parameters}
  \begin{tabular}{|c|c|}
    \hline
    Parameter & Tuning Range  \\
    \hline
    $\epsilon$ & 1e-300 - 0.9  \\
    \hline
    $\alpha$ & 1/5,2/5,3/5,4/5,1 \\
    \hline
    $l/n$ & 0.1,0.2,0.3,0.4,0.5,0.6,0.7,0.8,0.9  \\
    \hline
  \end{tabular}
\end{table}

% create hash sign

\subsection{Performance Metrics}\label{subsec2}

We evaluate the performance of the proposed method using the following two metrics:

\subsubsection*{Concordance}

Concordance \cite{gerds2013estimating} index is popular to evaluate the model performance in survival models/functional regression setup.  We define concordance index with right censored data.
The expression is as follows :
\begin{equation*}
  C^{td}(t) = Pr(\hat{S}_i(t_i) > \hat{S}_j(t_i) \mid d_i = 1 , t \leq t_j ,  t_i  <  t_j)
\end{equation*}

where $S_i(t_i)$ is the predicted survival probability of the $i^{th}$ individual at time $t_i$ and $S_j(t_i)$ is the predicted
survival probability of the $j^{th}$ individual at time $t_i$. The time-dependent
concordance index given above measure the level of agreement between the predicted survival
probabilities of two individuals and actual event time of the individuals. It is defined between
0.5 and 1, where 0.5 random prediction and 1 is perfect prediction, represents that every prediction 
is in agreement with the actual event time.

\subsubsection*{Integrated Brier Score}

 We use another benchmark as Integrated Brier Score.  Since the data consists of right censored observation we use the following modified version of the Integrated Brier Score, we call it as $IBS^{c}$.
 $IBS^{c}$ computed between time interval $[t_{1}, t_{\max}]$ as $$IBS^{c} = \frac{1}{t_{\max} - t_{1}}\int_{t_{1}}^{t_{\max}} BS^{c}(t) dt$$.  We estimate the integration numerically via trapezoidal rule.  
 Expression of Brier Score for the censored observation ($BS^{c}$) is :
  $$BS^{c}(t) = \frac{1}{N}\sum_{i = 1}^{N} \left[I(y_{i} \leq t, \delta_{i} = 1) \frac{(0 - \hat{S}(t \mid x^{i}))^{2}}{\hat{G}(y_{i})} + I(y_{i} > t)\frac{(1 - \hat{S}(t \mid x^{i}))^{2}}{\hat{G}(t)}\right]$$
where, $\hat{S}(t \mid x^{i})$ is the predicted conditional function and $\frac{1}{\hat{G}(t)}$ is the inverse probability of the censoring weight.  We assume $C > t$, where $C$ is the censoring point. 

\subsection*{Base Models}

The algorithm need base models to calculate the estimation on dataset $\dataset_l$. 
We have used the following base models:

\begin{enumerate}

\item \textbf{Survival Tree} : Survival Tree is a tree based method for survival analysis.   We split the data into different groups based on the log-rank test statistic during the construction of survival tree. The log-rank statistic tells us how different the survival curves of two groups are.  Further, we grow the tree
recursively, till it meets a certain stopping criteria.

\item \textbf{Random Survival Forest} : Random Survival Forest \cite{ishwaran2008random} is
 an extension Survival Tree, The output of Random Survival Forest is the average of the 
 output of Survival Tree, obtained from different bootstrap samples of the data.

\item \textbf{Cox Model (Lasso and Ridge)} : Cox Model is a semi-parametric model for survival analysis, which
is based on the proportional hazard assumption. The Cox
Model is given by

\begin{equation}
 h(t \mid x) = h_0(t) \exp(\beta^T x)
\end{equation}

where $h(t \mid x)$ is the hazard function of an individual with covariate $x$ at time $t$,
$h_0(t)$ is the baseline hazard function and $\beta$ is the parameter vector. The Cox Model
is usually estimated using partial likelihood method.  Regularisation brings two variations in the Cox Model, known as Lasso and Ridge Cox Model.  We use Breslow estimator \cite{breslow1972contribution} for calculation of the conditional survival function from Cox model.   

\item \textbf{K-Nearest Neighbour Survival} : K-Nearest Neighbour Survival ( \cite{chen2019nearest}) is a non-parametric method to estimate conditional survival function. In K-Nearest
Neighbour Regression, the prediction of the query point is the average of the K-Nearest
Neighbours of the query point. 
Suppose, we have the following right censored survival data, $\left( (X_1, Y_1, \delta_1), \cdots, (X_n, Y_n, \delta_n) \right) \in \mathcal{X} \times \mathbb{R}^{+} \times \{0, 1\}$ 
Then for specified Kernel function $K : \mathbb{R}^{+} \rightarrow \mathbb{R}^{+}$, and bandwidth $h > 0$, we can measure the proximity
between observation $X_i$ and $X_j$ by $K \left( \frac{\rho(X_i, X_j)}{h} \right) $, where $\rho$ is the distance function. We can create unique
event time, event count and survivor count as follows

\begin{equation*}
  \label{eventtime}
  \mathcal{Y}_{K}(x ; h) = \biggl\{Y_j : j \in [n] \cap \delta_jK\left(\frac{\rho(x, X_j)}{h}\right) > 0\biggr\}
\end{equation*}

\begin{equation*}
  \label{eventcount}
  \mathcal{D}_{K}(t \mid x; h) = \sum_{j=1}^{n}K\left(\frac{\rho(x, X_j)}{h}\right)\delta_j\mathbb{I}(Y_j = t)
\end{equation*}

\begin{equation*}
  \label{survivorcount}
  \mathcal{R}_{K}(t \mid x; h) = \sum_{j=1}^{n}K\left(\frac{\rho(x,X_j)}{h}\right)\mathbb{I}(Y_j \geq t)
\end{equation*}

Then the survival function is given by
\begin{equation}
  \label{kaplanmeier}
  \hat{S}_{K}(t \mid x; h) = \prod_{t^{'} \in \mathcal{Y}_{K}(x; h)} \biggl( 1 - \frac{\mathcal{D}_{K}(t^{'} \mid x; h)}{\mathcal{R}_{K}(t^{'} \mid x; h)} \biggr)^{\mathbb{I}(t^{'} \leq t)}
\end{equation}

The K-Nearest Neighbour Survival can be derived from the above equation by replacing the kernel function $K$ by the $\mathcal{I}$, where mathematical expression of $ \mathcal{I}(x; k, X_i) $\footnote{Taking $\mathcal{I}(\cdot) = 1 \forall i$  will lead to a population
Kaplan-Meier Estimate} is as follows :
\begin{equation}
  \label{knn}
  \mathcal{I}(x; k, X_i) = \mathbb{I} \biggl( \sum_{j=1, j \neq i}^{n} \mathbb{I}( \rho(x, X_i) \leq \rho(x, X_j) ) \geq k \biggr)
\end{equation}

\end{enumerate}

 KNN survival implementation is available by George Chen at \href{https://github.com/georgehc/npsurvival}{https://github.com/georgehc/npsurvival}.

\subsection*{Numerical Results on the Real Datasets}

Table \ref{tab:concordance} provides the information of the concordance of all the five models on the three datasets, whereas Table \ref{tab:ibs} describes the integrated brier score of all the five models.
We use 5 fold cross validation to obtain the result  We observe that our proposed construction of Combined regression strategy provides maximum concordance index and minimum integrated brier score for all datasets.  

%% create concordance table
\begin{table}[htbp]
  \centering
  \caption{Concordance (Higher is Better)}
  \label{tab:concordance}
  \begin{tabular}{|c|c|c|c|}
    \hline
    Dataset & METABRIC &  FLCHAIN &  RECID  \\
    \hline
    Survival Tree & 0.553 &  0.780 &  0.544 \\
    \hline
    Random Survival Forest & \dashuline{0.669} &  \dashuline{0.899} &  \dashuline{0.629} \\
    \hline
    Cox Model (Lasso) & 0.626 &  0.886 &  0.599 \\
    \hline
    Cox Model (Ridge) & 0.589 &  0.901 &  0.551 \\
    \hline
    KNN Survival Model & 0.561 &  0.829 &  0.540 \\
    \hline
    proposed & \textbf{0.677} &  \textbf{0.911} &  \textbf{0.630} \\
    \hline
  \end{tabular}
\end{table}

\begin{table}[htbp]
  \centering
  \caption{Integrated Brier Scoe (Lower is better)}
  \label{tab:ibs}
  \begin{tabular}{|c|c|c|c|}
    \hline
    Dataset & METABRIC &  FLCHAIN &  RECID  \\
    \hline
    Survival Tree & 0.278 &  0.065 &  0.262 \\
    \hline
    Random Survival Forest & \dashuline{0.167} &  \dashuline{0.046} &  \textbf{0.175}  \\
    \hline
    Cox Model (Lasso) & 0.160 &  0.079 &  \dashuline{0.183} \\
    \hline
    Cox Model (Ridge) & 0.162 &  0.143 & .189   \\
    \hline
    KNN Survival Model & 0.203 &  0.054 &  0.211 \\
    \hline
    proposed & \textbf{0.148} & \textbf{0.043} &  \textbf{0.175} \\
    \hline
  \end{tabular}
\end{table}

We also perform a D-calibration test \cite{haider2020effective} at 0.05 level for each of the 5-fold cross validation in all three datasets.  The Table-\ref{tab:Dcal} reports the number of times model passes D-calibration test.   We observe that proposed model passes the test maximum number of times than almost all weak learners.  However, it passes equal number of tests as the strongest weak learners Random Survival Forest.   The experiments validate not only the efficacy of the model over other models, but also confirms that the model's predicted probability calculations are meaningful.

 \begin{table}[ht]
    \centering
    \caption{Out of 5-fold number of times model passes D-calibration test at 0.05 level}
      \label{tab:Dcal}
    \begin{tabular}{|lcccc|}
      \hline
     Models & METABRIC &  FLCHAIN &  RECID & total \\
      \hline
      Survival Tree &   0 &      2 &     0 & 2 \\
      \hline
      Random Survival Forest &   5 &      4 &      5 & \dashuline{14} \\ 
      \hline
      Cox (lasso) &   4 &     3 &     5 & 12 \\
      \hline
      Cox (ridge) &   5 &     0 &      5 & \dashuline{10} \\
      \hline
      KNN Survival &   0 &      2 &      0 & 2 \\
      \hline
      proposed &   5 &      4 &      5 & \textbf{14} \\
       \hline
    \end{tabular}
  \end{table}

\section{Covariate Relevance}\label{sec5}

  The relevance of the covariates is also an important aspect of survival analysis. COBRA Survival
has an inbuilt mechanism to measure the relevance of the covariates. We use a logistic regression to find the relevance of the covariates
for a query point $x$. A detailed overview of
logistic regression is available at \cite{cramer2002origins}, on the indicator function \ref{indicator},
where $i = k+1, k+2, \cdots, n^{'}$.

The logistic regression is given by
\begin{eqnarray*}
  \label{logistic}
&&  \log \biggl( \frac{Pr(\Gamma_{i}(x;X_j,\mathcal{M},\epsilon,\alpha) = 1)}{1 - Pr(\Gamma_{i}(x;X_j,\mathcal{M},\epsilon,\alpha) = 1)} \biggr) \\ &= & \beta_0 + \beta_1 X_{1,i} + \beta_2 X_{2,i} + \dots + \beta_p X_{p,i} = \beta^T\mathbf{X}_i \\ && \forall i = k+1, k+2, \cdots, n^{'}\\
\end{eqnarray*}

The coefficients $\beta_1, \beta_2 \dots \beta_p$ are the relevance of the covariates $X_1, X_2, \cdots, X_p$ respectively.
We recall the definition of $\Gamma(\cdot)$ :
\begin{equation}
  % label this equation
  \label{indicator}
  \Gamma(x;X_j,\mathcal{M},\epsilon,\alpha) = \mathbb{I} \biggl( \sum_{m=1}^{\mid \mathcal{M} \mid} \mathbb{I}( d(S_{k,m}(t \mid x) , S_{k,m}(t \mid X_j))  \leq \epsilon) \geq \mid \mathcal{M} \mid\alpha \biggr)
\end{equation}

  The \textit{k} subscript notifies that the prediction of the base models takes training on the dataset $\dataset_k$.
Here, $\mathbb{I}$ is the indicator function, which is 1 if the condition is true and 0 if the condition is false. $\epsilon$ is
the threshold distance, $\alpha$ is thetakes fraction of base models in the consensus of $\epsilon$ proximity of the query point $x$.
basically $\mathcal{I}(.)$ is the indicator function, which is 1 if the $i^{th}$ individual is in the $(\epsilon , \alpha)-$proximity of the
query point $x$, otherwise 0.

\subsection*{Simulation Study}

  We use a synthetic population to demonstrate COBRA Survival's ability to discriminate between different 
factors and determine their relative importance. Our strategy incorporates a non-linear data-generating 
procedure supported by the following link rule\footnote{Similar Link rule is taken by \citeauthor{bellot2018boosted}}

\begin{equation*}
  \Lambda(\mathbf{X}_i) = 2 + \log(13X_{0,i} + 5X_{1,i} + 7X_{2,i}) + X_{3,i}
\end{equation*}

We have generated 2000 sample, where 40 percent of the sample are censored. The censoring time is generated
by taking uniform distribution between 0 and time of event. The time of event is generated by following
weibuill distribution 

\begin{equation*}
  T_i \sim \mathcal{W}(2, \Lambda(\mathbf{X}_i))
\end{equation*}

\begin{figure}[htbp]
  \centering
  \includegraphics[width=0.8\textwidth]{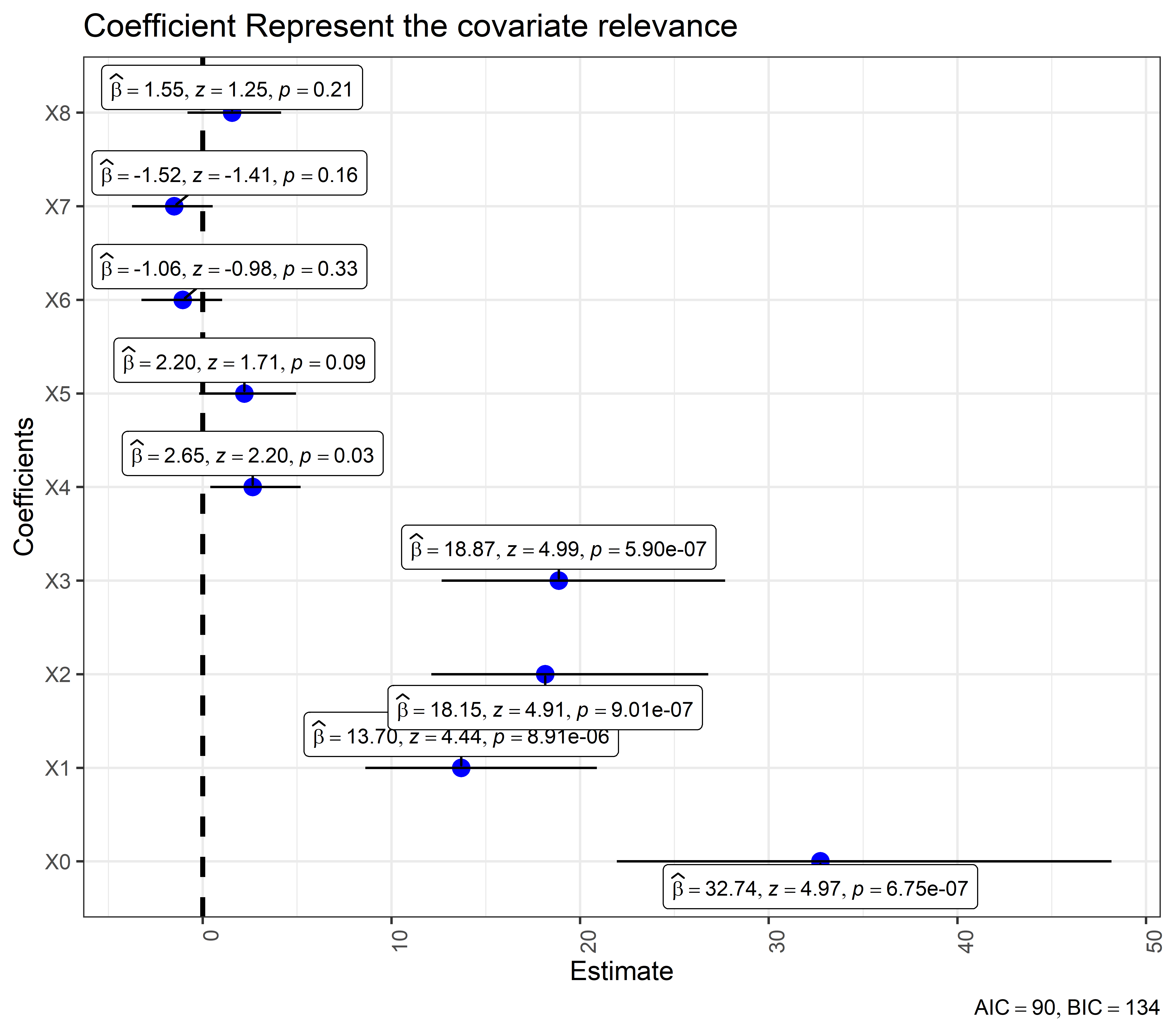}
  \caption{Relevance of the covariates}
  \label{fig:relevance}
\end{figure}

There are 9 covariates generated, where $X_0$ , $X_1$ , $X_2$ , $X_3$ have non-linear effect 
on true event time, whereas $X_4$ , $X_5$ , $X_6$ , $X_7$ , $X_8$ have no effect on true event time.
This is done to show that COBRA Survival can identify the covariates which have effect on true event time,
and the covariates which have no effect on true event time. The covariates are generated from uniform distribution.
The estimates of the covariate relevance is given in Figure \ref{fig:relevance}.

% #Add image relevance.png with label relevance

\section{Conclusion}\label{sec6}

 A large number of variations of cobra structure can not able to beat the performance of Random Survival Forest even with weak learners taken as Random Survival Forest. The performance of the Cobra depends on the appropriate choice of its parameters.  We observe that the proposed method in the paper works well for all three datasets.  Also, a lesser number of observations in proximity points affect the performance.  Our proposed model in the variable selection method works very well in the simulated framework. However, this requires special attention in a separate research. We may pursue further research on weighted version, problems on smaller observations, and more on variable importance in this context. The work is in progress.

\bibliography{sn-bibliography1}

\end{document}